\pdfoutput=1

\documentclass[11pt]{article}

\usepackage{acl}
\usepackage{booktabs} 
\usepackage{collcell}
\usepackage{etoolbox}
\usepackage{arydshln}
\usepackage{booktabs}
\usepackage{hyperref}
\usepackage{xcolor}
\usepackage{amsmath} 
\usepackage{colortbl}

\usepackage{times}
\usepackage{latexsym}
\usepackage{array}
\usepackage{amssymb}
\usepackage{graphicx}
\usepackage{subfigure}
\usepackage{romannum}
\usepackage{amsmath}
\usepackage{multirow}
\usepackage{algorithmic}
\usepackage{colortbl}
\usepackage{amssymb}
\usepackage{graphicx}
\newcolumntype{C}[1]{>{\centering\collectcell\stats}p{#1}<{\endcollectcell}}
\usepackage[ruled,linesnumbered]{algorithm2e}
\usepackage{longtable}
\usepackage{CJKutf8}

\newcommand{\stats}[1]{%
  \ifnum\pdfstrcmp{\unexpanded{#1}}{\unexpanded{---}}=0
    \textbf{#1}%
  \else
    \num{#1}%
  \fi
}
\usepackage{collcell}
\usepackage[T1]{fontenc}

\usepackage[utf8]{inputenc}

\usepackage{microtype}

\usepackage{inconsolata}

\title{Analyzing and Reducing Catastrophic Forgetting in Parameter Efficient Tuning}

\author{Weijieying Ren\textsuperscript{$1$}\hspace{0.5em}
Xinlong Li\textsuperscript{$2$} \hspace{0.5em}
Lei Wang\textsuperscript{$3$} \hspace{0.5em}
Tianxiang Zhao\textsuperscript{$1$}\hspace{0.5em}
Wei Qin\textsuperscript{$2$}  \\
\textsuperscript{$1$} College of Information Sciences and Technology, Pennsylvania State University \\
\textsuperscript{$2$}School of Computer Science and Technology, Singapore Management University\\
\textsuperscript{$3$}School of Computer Science, Hefei Univeristy of Technology\\
{\tt  \{wjr5337, tkz5084\}@psu.edu, 
lei.wang.2019@phdcs.smu.edu.sg,} \\
{ \tt qinwei.hfut@gmail.com}}

\begin{document}
\maketitle

\begin{abstract}
Existing research has shown that large language models (LLMs)
exhibit remarkable performance in language
understanding and generation. 
However, when LLMs are continuously fine-tuned on complex and diverse domain-specific downstream tasks, the inference performance on historical tasks decreases dramatically, which is known as a catastrophic forgetting problem.
A trade-off needs to be kept between learning plasticity and memory stability. Plenty of existing works have explored strategies like memory replay, regularization and parameter isolation, but little is known about the geometric connection of
various adjacent minima in the continual LLMs fine-tuning scenarios. 
In this work, we investigate the geometric connections of different minima through the lens of mode connectivity, which means different
minima can be connected by a low-loss valley. 
Through extensive experiments, we uncover the mode connectivity phenomenon in the LLMs continual learning scenario and find that it can strike a balance between plasticity and stability.
Building upon these findings, we propose a simple yet effective method called Interpolation-based LoRA (I-LoRA), which constructs a dual-memory experience replay framework based on LoRA parameter interpolations. 
Extensive experiments and analysis on eight domain-specific CL benchmarks demonstrate that
I-LoRA consistently
show significant improvement over the previous state-of-the-art approaches with up to $11\%$ performance gains, providing a strong baseline and insights for future research on the large language model continual learning problem.
Our code is available at \url{https://github.com/which47/LLMCL}.


\end{abstract}
\section{Introduction}


Despite the impressive zero-shot and few-shot learning capabilities demonstrated by generative Large Language Models (LLMs) \cite{wang2023comprehensive}, their performance degrades largely in the continual learning (CL) scenario, which requires the adaptation to complex new tasks while preserving previously learned knowledge \cite{razdaibiedina2023progressive}. This problem is known as catastrophic forgetting \cite{li2017learning}, highlighted by the trade-off between learning plasticity (fast adaptation to novel tasks) and memory stability (preservation of learned knowledge).

Previous works have explored three directions to achieve a balance between plasticity and stability \cite{wang2023comprehensive}.  
Among them, replay-based methods preserve historical information by explicitly storing a subset of historical data \cite{chaudhry2019tiny} or prompts \cite{khan2023introducing}.
Regularization-based methods \cite{li2017learning} propose to penalize change of important model parameters or conduct embedding-level distillation from the historical model during the learning of new tasks. 
The third category, parameter isolation methods \cite{kang2022forget}, mitigates forgetting by explicitly encoding task-specific model parameters, e.g., introducing a list of adaptors to consolidate historical knowledge.

Despite these explorations, we argue that the essence of achieving plasticity-stability trade-off lies in the intersection of loss landscapes surrounding optima for various tasks (see Figure 4). Once the model has converged to the optima of historical tasks, there would be a region around it that achieves optimal performance for the new task. Moreover, there exists a parametric path connecting the optima of historical tasks to that of the new task, and a well-balanced trade-off can be attained by traversing model parameters along this path. This phenomenon is named ``mode connectivity'' \cite{garipov2018loss,doan2023continual}. Similar observations are made in the computer vision domain, which unifies CL with multi-task learning \cite{mirzadeh2020linear} and conducts class incremental learning \cite{wen2023optimizing}. However, whether analogous observations hold in the LLM domain remains an unexplored task. LLMs often involve intricate pre-training on large-scale corpora, dealing with nuanced contextual understanding and text generation. Analyzing the mode connectivity phenomenon holds seamlessly to LLMs is crucial for facilitating continual learning on them.

In this paper, we are the first to present such a new perspective in understanding and improving CL for Large Language Models. 
Specifically, we focus on the following research questions in the context of Parameter-Efficient Fine-Tuning (PEFT) strategies: 
\textbf{RQ1}: \textit{Does {mode connectivity} exist for continual learning in PEFT?}
and \textbf{RQ2}:  \textit{How can we leverage the geometric connections of different minima to address catastrophic forgetting in PEFT?} 
Based on experiments conducted across eight diverse and domain-specific CL benchmarks, our analysis reveals the following findings: 1) mode connectivity exists in the continual fine-tuning scenario of LLMs; 2) mode connectivity can be established by constructing a parametric path to connect historical and current optima; and 3) a more optimal trade-off between stability and plasticity can be achieved along a linear trajectory between the two optima.
Based on these observations, we propose I-LoRA (Interpolation-based LoRA), which aims to simulate the weight interpolation process along with continual updates of LLMs using LoRA. Specifically, I-LoRA establishes a dual-memory framework by maintaining a fast learner parameterized by the working memory and a slow learner parameterized by the long-term memory. The fast learner is responsible for quick adapting to the evolving data, while the slow learner aims to consolidate the long-term memory and preserve historical knowledge. We iteratively update these two learners to achieve a balance between plasticity and stability.

The main contributions of this paper are listed as follows:
\begin{itemize}
    \item We are the first to present a novel perspective on understanding the continual learning for LLMs on mode connectivity;
    \item  Base on comprehensive analysis, we propose an effective CL algorithm for LLM by designing a dual-memory framework, with a fast learner to quickly adapt to evolving tasks and a slow learner to reduce forgetting;
    \item Extensive experiments and analysis of I-LoRA are conducted across diverse textual datasets, which validate its strong performance in the trade-off between plasticity and stability.
\end{itemize}

\section{Related Works}
\textbf{Continual Learning} focuses on sequential learning of non-stationary data, ideally accumulating previously gained knowledge \cite{wang2023comprehensive}. 
Based on the taxonomy in \cite{li2017learning}, 
existing works can be broadly classified into three dimensions: 
1)  Replay-based methodologies involve the reloading of historical raw data \cite{rolnick2019experience,buzzega2020dark} or the utilization of synthetic data \cite{lesort2019generative} generated from a generative model trained on historical data.
2) Regularization-based methods \cite{kirkpatrick2017overcoming,saha2021gradient,kim2023achieving} penalize model parameter change and balance the trade-off between plasticity and stability; 
3) Parameter isolation methods \cite{kang2022forget,golkar2019continual}
identify, allocate and incorporate critical parameters for different tasks during CL, thereby minimizing the interaction between tasks.
For an in-depth discussion on continual learning in the era of large language models, readers may refer to \cite{wang2023comprehensive}.

\begin{figure*}[htp]
    \centering
\includegraphics[width=1\linewidth]{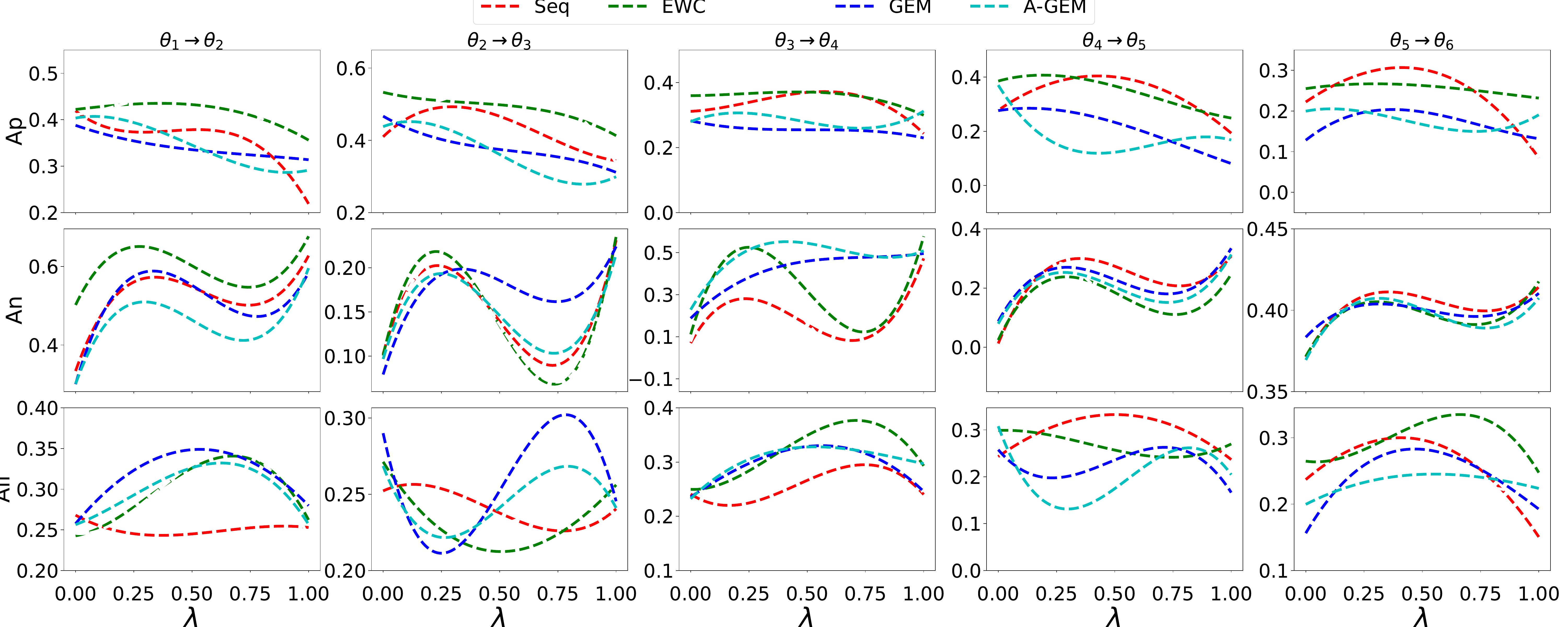}
    \caption{Inference accuracy curves along the linear connection between two adjacent continual minima of five representation continual learning baselines on seven domain-specific benchmarks. The y-axis, named Ap(upper row), An(mid row), and Aall(bottom row) denotes accuracy on previous tasks $1:t$, on the current task $t+1$, and on all learned tasks $1:(t+1)$ respectively. X-axis, $\lambda$ indicates the interpolation factor. Taking the testing accuracy as the measure of connectivity is because it is more sensitive to moving along the path
than training loss. }
    \label{fig:linear_connectivity}
\end{figure*}
\noindent \textbf{Linear Mode Connectivity} is a phenomenon that different minima can be connected by low-loss paths in the parameter space \cite{garipov2018loss,entezari2021role}.
Optimizing neural networks involves the finding of a minimum within a high-dimensional, non-convex objective landscape. 
\cite{frankle2020linear} asserts that, from the same initialization, local minima obtained with different training data orders can be interconnected by a linear low-loss path, thereby alleviating the challenge of curve identification. Building upon this discovery, recent research by \cite{mirzadeh2020linear}  observes that solutions in multitask and continual learning scenarios are connected by straightforward curves exhibiting low errors in weight space. 
This phenomenon, termed Linear Mode Connectivity, is empirically demonstrated to be a linear path when both multitask learning and continual learning share the same initialization weights.
However, these aforementioned works typically study mode connectivity using non-pretrained models in the field of computer vision \cite{wen2023optimizing,zhao2020bridging}, weight pruning analysis \cite{pellegrini2022neural}, loss landscape analysis \cite{frankle2020linear,garipov2018loss}, and etc.

Recently, \cite{qin2022exploring} pioneered the exploration of mode connectivity analysis on pretrained models, revealing that although minima trained on different tasks are inherently disconnected, pre-training gradually brings the optimal regions of diverse tasks closer in an implicit manner. In this study, we are the first to delve into continual learning within the framework of PEFT.

\section{Analyzing Linear Mode Connectivity in Parameter Efficient Continual Learning for LLMs}

In this section, we design an empirical study to answer RQ1, whether mode connectivity exists for continual learning in PEFT. We first introduce the notations and formulate the continual learning task before going into empirical details. 

CL can be formulated as learning from a sequentially ordered set of tasks $\{\mathcal{D}_1,\mathcal{D}_2,...,\mathcal{D}_T\}$, where each task is specified by input-label pairs. 
To be specific, the $t$-th task is specified by $\mathcal{D}_t = \{(\mathbf{x}_i,y_i)\}_{i=1}^{N_t}$, where $N_t$ represents the number of training examples for the $t$-th task.
Formally, the objective of CL is to learn a function $f:\mathbb{R}^d\rightarrow \mathbb{R}$
with parameters $\boldsymbol{\theta} \in \mathbb{R}^d$ that minimizes the loss over the  tasks:
\begin{equation}
\operatorname*{min}_{\boldsymbol{\theta}} \mathbb{E}_{t=1}^{T}[\mathbb{E}_{(x,y)\sim \mathcal{D}_{t}}[\ell(f_{\boldsymbol{\theta}}(\mathbf{x}),y)]],
\end{equation}
where $\ell$ is the learning objective of target tasks, e.g., cross-entropy loss. In this study, we adopt one representative PEFT approach, LoRA, and the model $f$ comprises a large amount of pre-trained fixed parameters and a small number of tunable parameters (the LoRA module). For the simplicity of annotation, throughout this paper, we use $\boldsymbol{\theta}$ to denote those tunable parameters.


\subsection{Mode Connectivity Evaluation}
Given the minima of two adjacent tasks, denoted as $\boldsymbol{\theta}_{t}$ and $\boldsymbol{\theta}_{t+1}$, we posit the existence of a continuous curve $\phi(\lambda): [0,1] \rightarrow \mathbb{R}^{\boldsymbol{\theta}}$ connecting these minima. This curve represents a trajectory in the parameter space that smoothly transits from $\boldsymbol{\theta}_{t}$ to $\boldsymbol{\theta}_{t+1}$. The linear path connecting the two minima can be expressed as follows:
\begin{equation}
    \phi(\lambda) = (1 - \lambda)\cdot \boldsymbol{\theta}_{t} + \lambda \cdot \boldsymbol{\theta}_{t+1}.
    \label{eq::1}
\end{equation}
Essentially, traversing along the curve described in Equation \ref{eq::1} allows for the evaluation of the interpolation performance between stability (where $\phi(0) = \boldsymbol{\theta}_{t}$) and plasticity (where $\phi(1) = \boldsymbol{\theta}_{t+1}$). It is expected that {a better trade-off would exist along this trajectory}, and the two endpoints $\boldsymbol{\theta}_{t-1}$ and $\boldsymbol{\theta}_{t}$ are smoothly connected without significant loss barrier or performance drop along the path.
The evaluation performance on historical tasks, current tasks, and all tasks are abbreviated as Ap, An, and All, respectively.

\textbf{Empirical Observations}
We investigate the existence of mode connectivity in Pre-trained Language Models (PLMs) during the continual fine-tuning across multiple downstream tasks in Figure 1. For space limitation, we report the performance on the first six tasks with the same task order in  Table 1. The sequence of tasks and experimental setups are further introduced in the experiment section, and we adopt LoRA during the tuning. After continually learned for each task, we conduct linear interpolations (parameterized by $\lambda$) between initial parameters (previous optima) and current ones (optima of the current task), and evaluate model performances with different $\lambda$ values.

From Figure 1, we obtain the following observations:
(1) The evaluation performance on the previous task $t$ (denoted as Ap) {could be significantly enhanced along the linear trajectory $\boldsymbol{\theta}_t \rightarrow \boldsymbol{\theta}_{t+1}$ compared to the initial point}. This result suggests that parameters obtained along this trajectory may replace $\boldsymbol{\theta}_t$ to achieve better memorization effects.
(2)
There are points along the linear path $\boldsymbol{\theta}_t \rightarrow \boldsymbol{\theta}_{t+1}$ yielding superior performance {w.r.t the averaged past and current tasks}, implying that a better trade-off between stability and plasticity can be achieved along this linear interpolation than the end-points.
(3) Moreover, there are even intervals along the linear path $\boldsymbol{\theta}_t \rightarrow \boldsymbol{\theta}_{t+1}$ that exhibit comparable or superior accuracy on the current task $t+1$ (denoted as An) when compared to both endpoints. This observation suggests that points sampled within such intervals may serve as a better checkpoint for the current task $t+1$, surpassing the efficacy of $\boldsymbol{\theta}_{t+1}$.
\section{Methodology}
\begin{figure*}[h]
    \centering
 \includegraphics[width=1\linewidth]{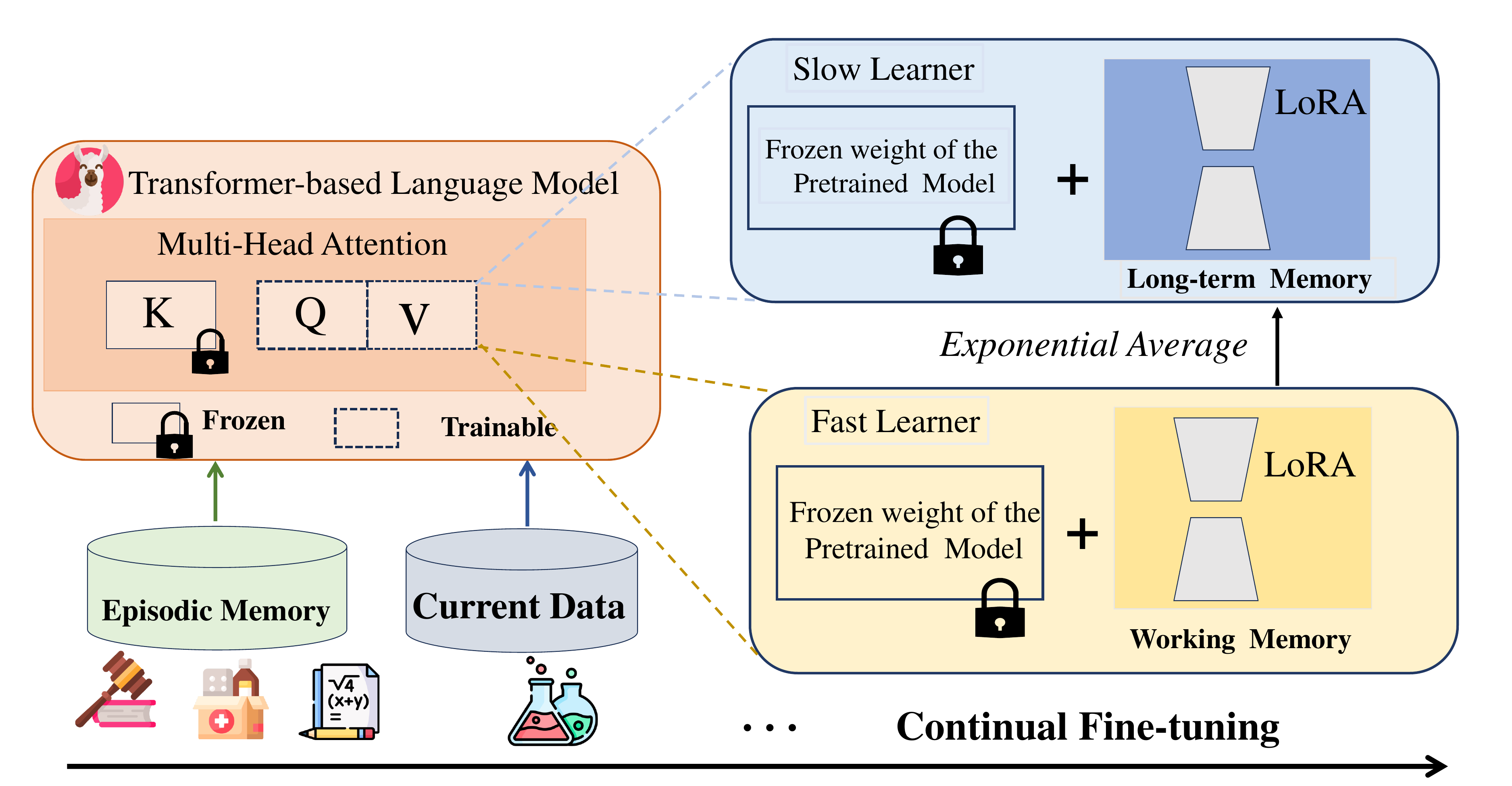}
    \caption{The framework of I-LoRA for Large Language Model Continual Learning. I-LoRA consists of a slow learner (depicted in blue) that learn long-term knowledge through exponential moving average of the fast learner weights; and (ii) a fast learner (depicted in yellow) retrieves historical knowledge while simultaneously adapting to current data. Both learners can be trained synchronously.}
    \label{fig:linear_connectivity}
\end{figure*}


Inspired by the existence of ``mode connectivity'', we propose a simple yet effective method, Interpolation-based LoRA (I-LoRA), to keep the balance between rapid adaptation and knowledge preservation in the PEFT process. I-LoRA constructs a dual-memory experience replay framework by maintaining a long-term memory $\boldsymbol{\theta}^l$ for stability and a working memory $\boldsymbol{\theta}^w$ for plasticity and improves the trade-off with the idea of interpolation across optima. Next, we will introduce the design of {dual memory} in Section 4.1 before presenting the full algorithm in Section 4.2.

\subsection{Dual Memory for Fast and Slow Learning}

In this work, we adopt a \textit{dual-memory} architecture to facilitate the separate encoding of historical and new optima, which enables us to explicitly estimate the trade-off. The framework comprises a fast learner (parameterized by working memory $\boldsymbol{\theta}^w$) and a slow learner (parameterized by long-term memory $\boldsymbol{\theta}^l$). The working memory, $\boldsymbol{\theta}^w$, is learned by simulating the fast learning of each new task. For task $t$, at each step $k$, it will be optimized on back-propagated gradients from modeling $\mathcal{D}_t = \{(\mathbf{x}_i,y_i)\}_{i=1}^{N_t}$. $\boldsymbol{\theta}^w$ can be understood as learning to arrive at the optima of this new task, converging to $\boldsymbol{\theta}^*_{t+1}$. 

To keep the balance between historical and new knowledge, we further leverage a long-term memory,  denoted as $\boldsymbol{\theta}^l$. As observed in Section 3.1, a better trade-off can often be discovered along the path connecting the optima $\boldsymbol{\theta}^*_{t-1}$ and $\boldsymbol{\theta}^*_{t}$. However, it is challenging and computationally extensive to explicitly identify the optimal $\lambda$ in Equation \ref{eq::1}. To mitigate this problem, we conduct an iterative update of $\boldsymbol{\theta}^l$ as an exponential moving average of the fast learner weights $\boldsymbol{\boldsymbol{\theta}^w}$ along the trajectory:
\begin{equation}
  \boldsymbol{\boldsymbol{\theta}}^l_k = \lambda \cdot \boldsymbol{\boldsymbol{\theta}}^l_{k-1} + (1 - \lambda) \cdot \boldsymbol{\boldsymbol{\theta}}^w_k, 
\end{equation}
in which the step size $\lambda$ is a fixed hyper-parameter and $k$ denotes the update step. At each step, the previously learned $ \boldsymbol{\boldsymbol{\theta}}^l_{k-1}$ will also modulate the obtention of $ \boldsymbol{\boldsymbol{\theta}}^l_{k-1}$ to encourage a data-driven tuning of memorization effect, the detail of which will be introduced next.
\begin{samepage}
\begin{algorithm}
  \caption{I-LoRA} \label{alg:algo}
  \footnotesize
  \small
  \raggedright
  Input data stream $\mathcal{D}$, memory $\mathcal{M}$, Learning rate $\eta$, update frequency $a$, update ratio $\lambda$ \\
  \For {$t \in [1, 2, \dots, T]$}{
    \For {$k$ \text{in} Training steps}{
      Sampling $(\mathbf{x}, y) \in \mathcal{D}_t \bigcup \mathcal{M}$ \\ 
      $\mathcal{L}_{CE} \leftarrow \text{cross-entropy}(f(\mathbf{x};\boldsymbol{\theta}^{w}), y)$ \\
      Sampling $(\mathbf{x}_m, y_m) \in \mathcal{M}$ \\
      $\mathbf{z}_m \leftarrow f_o(\mathbf{x}_m; \boldsymbol{\theta}^l_k)$ \\
      $\mathcal{L}_{MSE} \leftarrow \text{mean-square}(f_o(\mathbf{x}_{m};\boldsymbol{\theta}^{w}), \mathbf{z}_m) $ \\
          $\mathcal{L} =  \mathcal{L}_{CE}  + \gamma \cdot \mathcal{L}_{MSE}$\\
      $\boldsymbol{\theta}^w_k \leftarrow \boldsymbol{\theta}^w_{k-1} - \eta \nabla_{\boldsymbol{\theta}^w_{k-1}} \mathcal{L}$\\
    $\boldsymbol{\theta}^l_k \leftarrow \lambda\boldsymbol{\theta}^l_{k-1} + (1 - \lambda)\boldsymbol{\theta}^w_k$ \\
    }
    $\mathcal{M} \leftarrow \{\mathbf{x}_t, y_t\} \bigcup \mathcal{M}$
  }
\end{algorithm}
\end{samepage}

\subsection{Continual PEFT with Dual Memory}

Now we present details of the proposed continual PEFT algorithm, which is summarized in Algorithm 1. Both the working memory $\boldsymbol{\theta}^w$ and long-term memory $\boldsymbol{\theta}^l$ are implemented as LoRA modules. And we adopt the classical experience replay (ER) as our backbone framework: a subset of historical data is kept in an external episodic storage $\mathbf{x}_m \in \mathcal{M}$ , which will be mixed with the current dataset $\mathbf{x}_t \in \mathcal{D}_t$ during learning. 

For task $t$, {during each training step, we optimize the fast learner} (parameterized by $\boldsymbol{\theta}^w$) using (1) the classification objective, as in line $5$ of Algorithm 1, and (2) the deviation of historical instance embeddings compared to the slow learner (parameterized by $\boldsymbol{\theta}^l$), as in line $8$. The former objective is implemented as a cross-entropy loss on the mixed data $\mathcal{M} \bigcup \mathcal{D}_t$,  while the latter objective is implemented as the MSE loss on embeddings:
\begin{equation}
\begin{aligned}
 \min_{\boldsymbol{\theta}^w}\mathcal{L} = & \mathbb{E}_{\mathbf{x} \in \mathcal{M} \bigcup \mathcal{D}_t} \mathcal{L}_{CE}(f(\mathbf{x}; \boldsymbol{\theta}^w))  \\
 & + {\gamma} \cdot \mathbb{E}_{\mathbf{x} \in \mathcal{M}} \mathcal{L}_{MSE}(f_o(\mathbf{x};\boldsymbol{\theta}^w); \mathbf{z}),
\end{aligned}
\end{equation}
where we omit task index $t$ for simplicity. In this equation, $f_o$ denotes the embedding extractor part of $f$, which maps the input into a representation space. $\mathbf{z}$ records the embeddings generated by the slow learner, $f_o(\mathbf{x};\boldsymbol{\theta}^w)$ represents the output of the fast learner, and $\gamma$ is a hyper-parameter controlling the weight of embedding deviation loss. An update of $\boldsymbol{\theta}^w$ is provided in Line $10$ of Algorithm 1. After each step, the slow learner will be updated 
as an exponential moving average of the fast learner
weights,
as in line $11$.

\section{Experiments}
\subsection{Experiment Setup}
\subsubsection{Dataset Description}
To undertake a critical assessment of the 'adaptation' and 'forgetting' capabilities of LLMs \cite{wang2023trace,gao2023unified}, we construct the dataset under three key considerations:
(I) \textit{Domain Specificity}, avoiding prior exposure to the majority of LLMs;
(II) \textit{Diversity}, the dataset should be diverse and complex w.r.t corpus format, linguistic aspects, and reasoning challenges;
(III)
\textit{Common sense knowledge}, In realistic scenarios, we aim to maintain the common sense knowledge in LLMs even when continually fine-tuning on diverse downstream tasks.
To achieve such goals, we organize two types of datasets:

\textbf{CL benchmarks for LLMs} is to satisfy goal (I) and (II). Specifically, we organize datasets to demonstrate the following properties: 
1) \textit{Domain Specificity}. 
Continual training on diverse downstream tasks signifies a realistic and promising avenue for advancing LLMs. We select dataset from education domain, i.e., ScienseQA \cite{lu2022learn}, clinical domain i.e., MedMCQA \cite{pmlr-v174-pal22a}, financial domain, i.e.,  FOMC \cite{shah-etal-2023-trillion}, legal domain, i.e., JEC-QA \cite{zhong2019jec},  and political domain. i.e.,  MeetingBank \cite{hu2023meetingbank}. 
2) \textit{Multilinguality}.
Cross-lingual Continual Learning poses a formidable challenge for LLMs attributed to vocabulary discrepancies and variations in pre-training corpus. Following \cite{wang2023trace},  We select C-STANCE \cite{zhao2023c} and 20Minuten \cite{kew202320} as multi-lingual dataset. 
3) \textit{Mathematical reasoning}. Mathematical problems involve complex logical operations, providing a test-bed for the reasoning ability of LLMs. Here, we leverage the popular NumGLUE dataset \cite{mishra2022numglue}. 

\textbf{General benchmarks for LLMs} is to satisfy goal (III). We adopt MMLU \cite{hendrycks2021measuring}, BBH \cite{suzgun2022challenging}, and PIQA \cite{bisk2019piqa} as our benchmark. A detailed description can refer to Appendix.

\subsubsection{Metric}
Let $R_{i,j}$ represents the inference accuracy on $j$-th task after training on the $i$-th, we evaluated the inference performance by averaging accuracy after the
training of on the $t$-th task as $Acc_t$:
\begin{equation}
Acc_t = \frac{1}{T}\sum_{i=1}^{t}R_{t,i}.
\end{equation}
Besides, we evaluate the memorization ability by
evaluating the backward transfer ability ($BWT$) that averages influence of learning the $t$-th task on all old tasks as $BWT_t$ \cite{wang2023comprehensive}:
\begin{equation}
 BWT_t = \frac{1}{t-1}\sum_{j=1}^{t-1}(R_{t,j}-R_{j,j}).
\end{equation}
where $T$ is the total number of data sequences.

\subsubsection{Baselines}
\begin{table*}[h]\centering 
\renewcommand{\arraystretch}{1.2}
\resizebox{\textwidth}{!}{
\begin{tabular}{lcccccccc}\toprule
\multicolumn{9}{c}{\textbf{Doman-specific CL benchmarks for LLMs}} \\
\hline
&C-STANCE &FOMC &MeetingBank &ScienceQA &NumGLUE-cm &20Minuten &MedMCQA &JEC-QA \\
\cline{2-9}
Seq-Train &41.8 &41.1 &31.2 &27.4 &21.9 &13.6 &25.6 &21.0 \\
ER &40.8 &45.6 &30.6 &29.4 &18.7 &16.5 &22.8 &22.4 \\
EWC &42.2 &53.0 &35.9 &38.5 &25.5&26.2 &23.5 &22.5 \\
GEM &38.8 &46.4 &28.3 &27.4 &12.7 &17.7 &28.5 &20.4 \\
A-GEM &40.2 &43.9 &28.0 &36.9 &19.8 &22.6 &27.3 &29.6 \\
L2P &43.8 &39.5 &24.0 &26.4 &22.7 &15.4 &24.6 &25.6 \\
PP &37.2 &42.1 &26.5 &28.3 &25.3 &25.9 &26.2 &21.1 \\
\hline
I-LoRA &44.4 &53.9 &30.6 &40.1 &33.7&27.3 &38.1 &36.3 \\
\bottomrule
\end{tabular}
}
\caption{Summary of the results on Nine domain-specific CL benchmarks with the Llama-2-7B. Averaged inference accuracy on the downstream tasks ($Acc_t$) is reported.}\label{tab:acc}
\end{table*}

\begin{table*}[h]\centering 
\renewcommand{\arraystretch}{1.2}
\resizebox{\textwidth}{!}{
\begin{tabular}{lcccccccc}\toprule
\multicolumn{9}{c}{\textbf{Downstream CL benchmarks for LLMs}} \\
\hline
&C-STANCE &FOMC &MeetingBank &ScienceQA &NumGLUE-cm &20Minuten &MedMCQA &JEC-QA \\
\cline{2-9}
Seq-Train &0&-10.0 &-11.1 &-13.4 &-17.5 &-26.0 &-12.6 &-16.1 \\
ER &0 &-7.0 &-12.9 &-15.2 &-24.4 &-26.2 &-18.2 &-14.8 \\
EWC &0 &-3.3 &-10.2&-12.2 &-20.6 &-19.1 &-19.4 &-18.5 \\
GEM &0 &-3.7 &-12.8 &-13.6&-24.5 &-21.8 &-9.5 &-17.6 \\
A-GEM &0 &-5.6 &-13.0 &-7.3 &-22.0 &-19.1 &-12.4 &-8.3 \\
L2P &0 &-9.0 &-16.5 &-11.8 &-13.4 &-21.4&-10.6&-8.6 \\
PP &0 &-2.4&-10.1&-9.7&-10.6&-10.8 &-9.0 &-14.2 \\
\hline
I-LoRA &0 &-0.6 &-12.7 &-6.1 &-9.1&-15.2 &-2.2 &-2.3\\
\bottomrule
\end{tabular}
}
\caption{Summary of the results on Nine domain-specific CL benchmarks with the Llama-2-7B. Averaged memorization performance ($BWT_t$) is reported.}\label{tab: }
\end{table*}

\begin{figure*}[th]
  \begin{subfigure}{}
\includegraphics[width=0.35\textwidth]{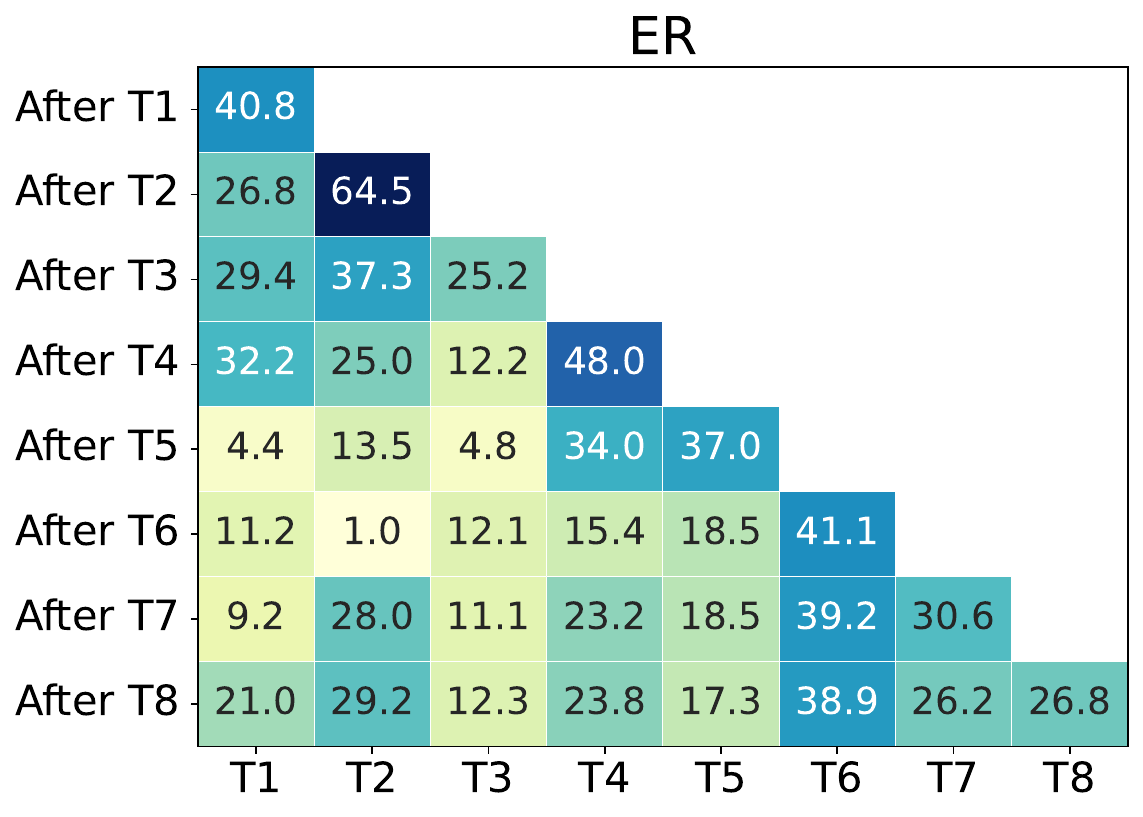}
\end{subfigure}
\begin{subfigure}{}
\includegraphics[width=0.3\textwidth]{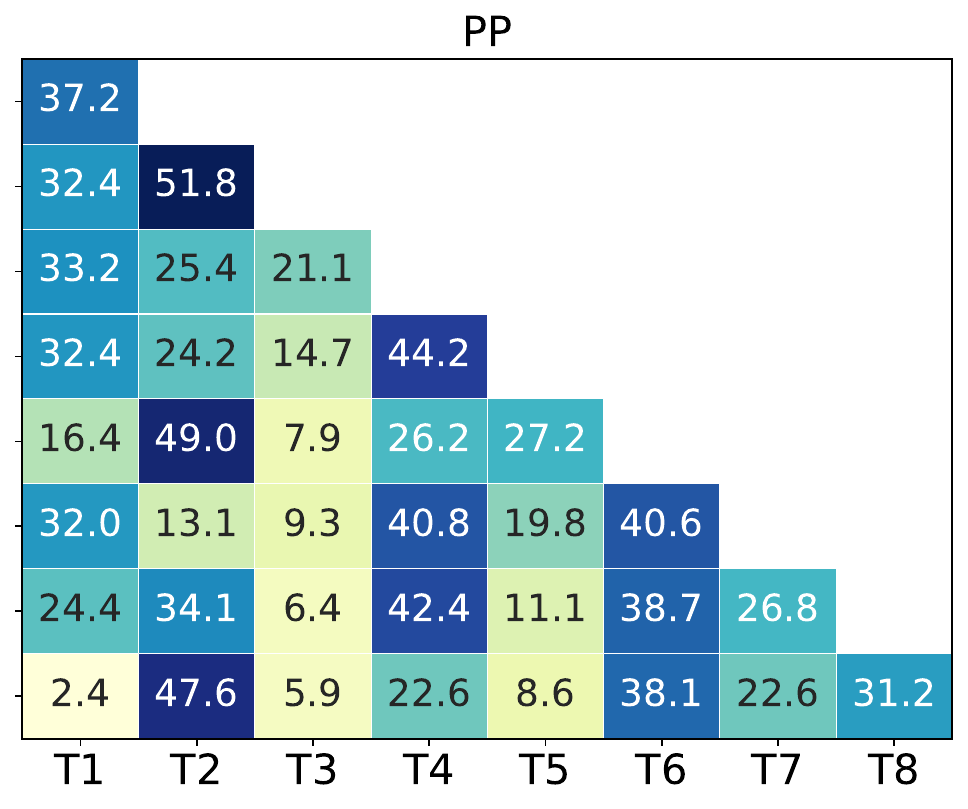}
\end{subfigure}
\begin{subfigure}{}
\includegraphics[width=0.3\textwidth]{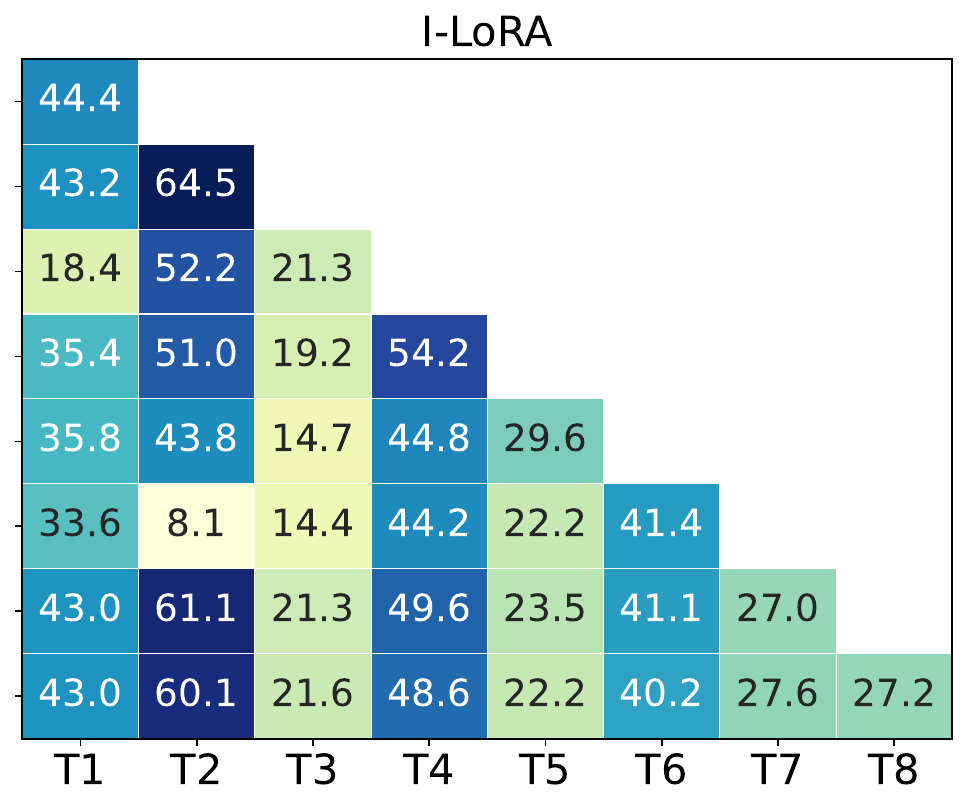}
\caption{Task-wise performance of CL methods when Llama-2-7B is continually fine-tuned on the sequential tasks. The heatmaps provide the test set of each task (x-axis) evaluated at the
end of each sequential learning task (y-axis).}
\end{subfigure}
\label{Fig:task_wise}
\end{figure*}
We evaluate the performance of I-LoRA against nine representative baseline methods. All baselines adopt the same architecture with LoRA for efficient fine-tuning. We maintain consistent parameter configurations across both I-LoRA and its comparative methods, thereby ensuring a fair comparison.
We use the following baselines: (1) Zero-shot inference (ZSI): zero-shot inference on sequential downstream tasks directly without tuning model parameters or adding prompts.
(2) Sequence Fine-tuning (Seq-Train): continual tuning all model parameters on a sequence of tasks (without adding any regularization or replaying samples from the previous tasks).
(3) Experience Replay (ER) \cite{chaudhry2019tiny} typically
stores a few old training samples within a small
memory buffer. 
(4) Elastic Weight Consolidation  (EWC) \cite{kirkpatrick2017overcoming} constrains important parameters to stay close to their old values and leverages the Fisher information matrix to measure the distance.
(5) Gradient Gradient Episode Memory (GEM) \cite{saha2021gradient}: maintains the gradient subspace important to the old tasks (i.e., the bases of core gradient space) for orthogonal projection in updating parameters.
(6) Average Gradient Episode Memory (A-GEM) \cite{chaudhry2018efficient}: a simple version of GPM that only stores the average gradient matrix on historical data. 
(7) Learning to Prompt (L2P) \cite{wang2022learning}: uses the input to dynamically select and update prompts from the prompt pool in an instance-wise fashion. 
(8) Progressive Prompt (PP) \cite{razdaibiedina2023progressive}: learns a new soft prompt for each task and sequentially concatenates it with the previously learned prompts, while keeping the base model frozen.
(9) Multi-task Learning (MTL): train a model on all tasks as multi-task
learning. This method is assumed as the upper bound of continual learning \cite{wang2023comprehensive}.

\subsubsection{Implementation Details}
Experiments are conducted on two RTX 4090 GPUs. We adopt Llama-2-7B as the foundational model and fine-tune LoRA \cite{hu2021lora} for continual learning purposes. The learning rate is set as 1$e$-4, accompanied by a linear warmup ratio of 0.2. Following \cite{wang2023trace}, we leverage the HuggingFace Transformers \cite{wolf-etal-2020-transformers} library for experiment implementation. Regarding the LoRA hyper-parameters, 
r is set to 8, and LoRA is integrated into the query and value matrices, with the LoRA alpha parameter configured to 16.

For each dataset, we curate a training set comprising 5000 samples and an evaluation set comprising 500 samples. Batch-size is set as 16. Notably, in the case of MedMCQA and JEC-QA, our sampling process exclusively focuses on single-choice questions. Detailed demonstrates about prompts are shown in Table \ref{tab:dataset-prompts} in the Appendix.

\subsection{Overall Comparison in CL}
We first conducted a comparative analysis of our approach against representative CL baselines from two dimensions. First, we assessed the adaptation and memorization capabilities during continual learning on domain-specific CL benchmarks. Second, we evaluated the memorization ability of general knowledge when continually fine-tuning on domain-specific CL benchmarks.

\textbf{Generalization Ability Assessment on Domain-specific CL benchmarks.} 
We start with a fine-tuned LLaMA-7B language model on each domain-specific CL benchmark, then test the $Acc_t$ performance to evaluate the adaptation performance. 
From Table 1, we observe that: 1) 
starting from a fine-tuned LLaMA-7B language model, CL minima on different tasks can be connected by a low-loss valley, and ensembling over the valley shows improved performance and generalization ability.
It is obvious that our approach, I-LoRA, consistently outperforms previous methods and shows a remarkable improvement (i.e., ranging from 3$\%$ to 10$\%$ accuracy gains) over the previous state-of-the-art CL methods.
2) 
I-LoRA consistently demonstrates superiority with an increasing number of historical tasks.
This observation suggests that leveraging mode connectivity in LLMs could enhance long-term memorization ability and validate the effectiveness of long-term memory in I-LoRA.

\textbf{Memorization Ability Assessment on Domain-specific CL benchmarks.} 
In this part, we explore the memorization capability of continual learning (CL) methods, specifically examining the extent to which these methods can mitigate the issue of catastrophic forgetting. From Table 2, we can make the following observations: 1) I-LoRA exhibits superiority in mitigating forgetting issues and demonstrates remarkable memorization ability. This observation validates our motivation and methodology design. I-LoRA adjusts parameters relying on the interpolation of mode connectivity, and its performance remains relatively stable throughout continual learning processes. 2) Existing CL-based methods exhibit weak performances when facing complex memorization tasks, such as those with high domain diversity and multilingualism. For example, one popular CL algorithm, EWC, shows a forgetting performance of $20.6\%$ and $19.1\%$ after fine-tuning on the mathematical NumGLUE-cm and German-based 20Minute dataset respectively. The diversity of sequential tasks makes these approaches ineffective.
In contrast, our method achieves consistently promising performance, e.g., I-LoRA decreases the forgetting score to $9.1\%$ on the NumGLUE-cm dataset. 
This boost in performance further validates our insight, which involves interpolating between adjacent minima and traversing along this path.

\textbf{Fine-grained Analysis on Task-wise Performance}
To better understand how various methods achieve a balance between stability and plasticity, we analyze how task-wise performance evolves as the model learns tasks sequentially. Experimental results are shown in Figure 3.
The diagonal of the heatmap demonstrates the plasticity of the model as it denotes the learning of each new task. Due to the limit of space, we select two representative methods ER and PP as the baseline.
The full experimental results are shown in Appendix.

Figure 3 demonstrates that the proposed I-LoRA offers a more consistent performance across the sequentially learned eight tasks compared to the baselines, showcasing a commendable balance between stability and plasticity. 
For example, the inference performance of ER on T3 dataset decreases from $12.2\%$ to $4.8\%$ at the endpoint of fine-tuning on T4 and T5, respectively. 
Similarly, the performance of PP drops from $14.7\%$ to $7.9\%$. 
On the contrary, the proposed I-LoRA demonstrates a good trade-off between stability and plasticity, decreasing from $19.2\%$ to $14.7\%$. 
After fine-tuning on T5, both ER and PP exhibit a much lower inference accuracy on previous tasks. For instance, ER achieves accuracy of $4.4\%$, $13.5\%$, and $4.8\%$ on T1, T2, and T3, respectively. In contrast, I-LoRA demonstrates superior stability, achieving accuracy of $35.8\%$, $43.8\%$, and $14.7\%$ on them.

Overall, I-LoRA provides an effective approach to leverage mode connectivity in continual fine-tuning of LLaMA-7B, enabling better utilization of long-term memory. This facilitates the effective consolidation of information across tasks and further mitigates forgetting.

\begin{table}[!htp]\centering
\renewcommand{\arraystretch}{1.2}
\caption{Performance on General Benchmarks after Fine-Tuning on Domain-Specific CL Benchmarks.}\label{gen loss}
\small
\centering
\begin{tabular}{lccccc}\toprule
\textbf{Method} &\textbf{MMLU} &\textbf{BBH} &\textbf{PIQA} &\textbf{GEN loss}\\\midrule
Zero-Shot &46.8  &38.2  &78.3   & ----- \\
\hline

Seq   &3.68  &28.82 &58.49  & -24.1\\
 ER    &5.22  &28.67 &53.1   & -25.41\\
EWC   &14.27 &34.18 &51.85  & -21.0\\
GEM   &15.45 &31.74 &53.48  & -20.88\\
A-GEM  &6.46  &32.44 &53.92  & -23.49\\
L2P   &2.24  &31.95 &54.19  & -24.97\\
PP    &30.58 &16.97 &53.05  & -20.9\\
I-LoRA  &15.77 &32.66 &51.25  & -21.21\\
\hline
MTL   &13.97 &31.92 &52.99  & -21.47\\
\bottomrule
\end{tabular}
\end{table}
\textbf{Performance on General Tasks}
Evaluating the performance on general tasks is important in evaluating the memorization and reasoning abilities of Large Language Models (LLMs) after fine-tuning on domain-specific tasks.
Table 3 displays the performance of continual learning (CL) methods, zero-shot inference performance of LLAMA-7B (zero-shot), and multi-task learning method (MTL). 
Detailed evaluation metrics and dataset descriptions can be referred to Appendix.
\begin{figure*}[h]
  \begin{subfigure}{}
\includegraphics[width=0.31\textwidth]{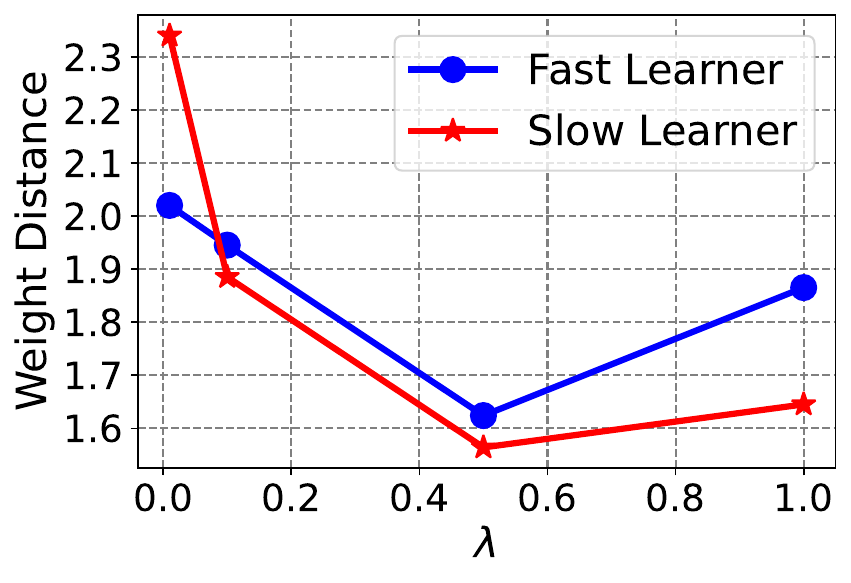}
\end{subfigure}
\begin{subfigure}{}
\includegraphics[width=0.31\textwidth]{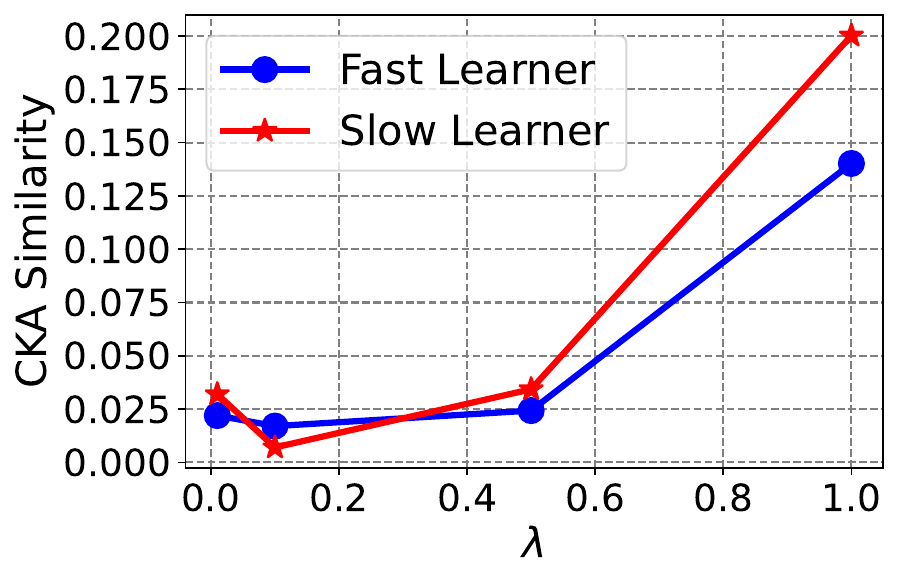}
\end{subfigure}
\begin{subfigure}{}
\includegraphics[width=0.31\textwidth]{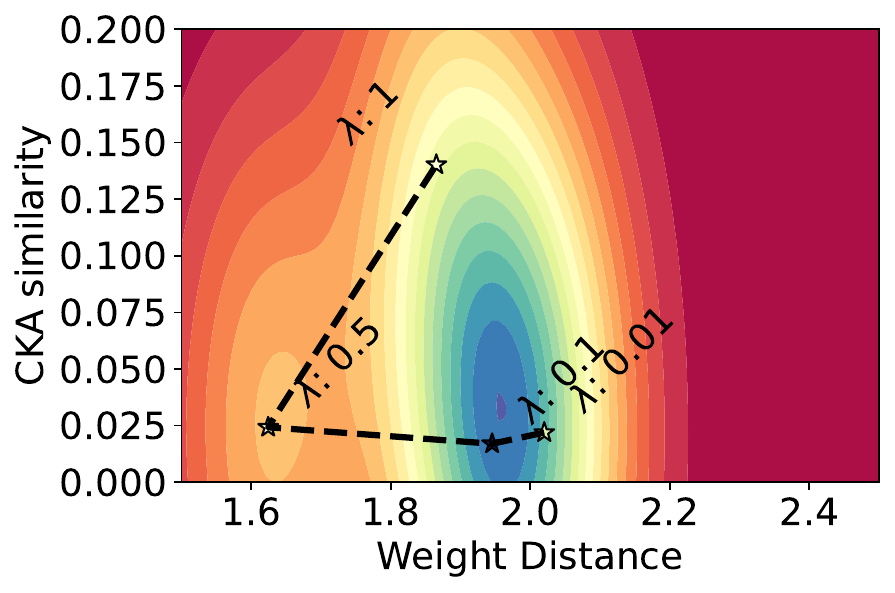}
\caption{Task-wise performance of CL methods when Llama-2-7B is continually fine-tuned on the sequential tasks. The heatmaps provide the test set of each task (x-axis) evaluated at the
end of each sequential learning task (y-axis).}
\end{subfigure}
\label{Fig:task_wise}
\end{figure*}
From these results, it can be observed that after continual learning processes, I-LoRA can still achieve an on-par performance with most baselines in these general language modeling tasks, despite a significant improvement in those specialized text domains as shown in Table 1 and 2. This phenomenon validates the advantage of I-LoRA in improving CL performance of LLMs. 

\subsection{Discussion}
To further evaluate the effectiveness of I-LoRA in the continual refinement of LLaMA-7B, as discussed in Section 3, we examine how I-LoRA achieves a balance between plasticity and stability from three perspectives: 1) Weight Distance; 2) Centered Kernel Alignment; and 3) Mean Accuracy Landscape.

\textbf{Weight Distance}
One intuitive explanation w.r.t the problem of catastrophic forgetting posits that after adapting to new data, LoRA parameters would change and converge toward another local optima, which deviates from the historical one.
Consequently, under the isotropic assumption of loss landscapes, the distance between historical and new weights can be used as a proxy for estimating the memorization of models. If parameters exhibit minimal change, it is rational to anticipate a lesser degree of forgetting. To this end, we propose to adopt the weight distance metric:
\begin{equation}
WD_l = ||\boldsymbol{\theta}_t^l - \boldsymbol{\theta}_{t+1}^l||_2,
\end{equation}
\begin{equation}
    WD_w = ||\boldsymbol{\theta}_t^w -
\boldsymbol{\theta}_{t+1}^w||_2.
\end{equation}

To evaluate the impact of weight interpolation, we measure the weight distance when varying $\lambda$ to different values, and visualize the analysis results in Figure 4. 
When $\lambda = 0$, the effect of I-LoRA is similar to ER.
Due to the constrained retention of memory samples, the current parameters of the fast learner, $\boldsymbol{\theta^s}_{t+1}$ , may diverge significantly from its previous counterpart, $\boldsymbol{\theta^s}_{t}$, resulting in a substantial weight distance.
Hence, as the value of $\lambda$ increases, the current model weights tend to approach the previous weights more closely. However, the weight distance increases as $\lambda$ is further raised. One plausible explanation is that the loss landscape becomes flatter in neighboring regions, and higher interpolation values may push the minima beyond these flat regions. Further analysis is provided in the Mean Accuracy Landscape Visualization section.

\textbf{Centered Kernel Alignment}
In addition to considering weight distance, we further examine the produced representation space. To this end, we utilize Centered Kernel Alignment (CKA) \cite{kim2023achieving,mirzadeh2020linear} to assess the similarity of LoRA's output representations.
A higher similarity score indicates a greater stability and memorization ability of the continual minima. 

The middle row of Figure 4 shows the CKA similarities with different interpolation ratios $\lambda$ of the working memory and long-term memory. It is obvious that the similarity in feature representation increases with a higher number of $\lambda$. Essentially, a higher $\lambda$ indicates a slower update process based on historical LoRA parameters, which contributes to the stability of LLMs. 

\textbf{Embedding Landscape Visualization}
To illustrate the geometric characteristics of the landscape across different continual minima, as described in Figure 4, we depict the landscape of embedding changes after learning on task 1 and task 2 by perturbing LoRA parameters. Concretely, we vary the parameter on the subspace constructed by $\boldsymbol{\theta}^w$ and $\boldsymbol{\theta}^l$, and visualize the extent of embedding change under different parameter interpolations.

As depicted in Figure 4, it's evident that $\lambda$ regulates the interpolation effects between continual minima and influences the convergence position on the loss landscape. A small value of $lambda$ encourages adjacent minima to remain close, while an increasing value of $lambda$ promotes slight changes in LoRA parameters and demonstrates high representation similarity. On the other hand, when the converged minima significantly diverges from the neighboring area of previous minima, LoRA will lose its capability in the trade-off.


\section{Conclusion}

Our empirical analysis provides comprehensive validation of the existence of intersections in loss landscapes surrounding task optima during Parameter-Efficient Fine-Tuning (PEFT) for LLMs. Building on this insight, we introduce I-LoRA, a pioneering approach that leverages two independent modules functioning as fast and slow learners, respectively. By promoting convergence between these modules and employing a linear interpolation, I-LoRA achieves a nuanced trade-off between plasticity and stability. As far as we are concerned, I-LoRA pioneers in enhancing CL for LLMs, and provide further opportunities for future explorations

\section{Limitations}
Although we believe that leveraging mode connectivity could potentially balance the trade-off between stability and plasticity, further theoretical analysis is warranted, and we have not yet explored this aspect in this study. In future research, we aim to develop a more comprehensive method that connects mode connectivity to continual fine-tuning for LLMs.


\bibliography{custom}

\appendix
\begin{table*}[t]
    \centering
    \footnotesize
    \begin{tabular}{l|p{0.6\linewidth}}
        \toprule
        \textbf{Datasets} & \textbf{Prompts} \\
        \midrule
        C-STANCE & \begin{CJK*}{UTF8}{gbsn}判断以下文本对指定对象的态度，选择一项：A.支持，B.反对，C.中立。输出A，B或者C。\end{CJK*} \\
        \midrule
        FOMC & What is the monetary policy stance for the following text? A. dovish, B. hawkish, C. neutral. Choose one from A, B, and C. \\
        \midrule
        MeetingBank & Write a summary of the following meeting transcripts. \\
        \midrule
        ScienceQA & Choose an answer for the following question and give your reasons.\\
        \midrule
        NumGLUE & Solve the following math problem.\\
        \midrule
        20Minuten & Provide a simplified version of the following paragraph in German. \\
        \midrule
        MedMCQA & Solve the following medical problem by choosing the correct answer from the following four choices.\\
        \midrule
        JEC-QA & \begin{CJK*}{UTF8}{gbsn}根据以下法律问题，从选项A，B，C，D中选择一项正确的答案\end{CJK*} \\
        \bottomrule
        \end{tabular}
    \caption{Datasets' Prompts}
    \label{tab:dataset-prompts}
\end{table*}
\section{Detail Results of Performance in Domains Specific Benchmarks and General Benchmarks}
\label{detail results of general benchmarks}
We utilize OpenCompass \cite{2023opencompass} with default settings to evaluate performance across three general benchmarks: MMLU \cite{hendrycks2021measuring}, BBH \cite{suzgun2022challenging}, and PIQA \cite{bisk2019piqa}.
Some detailed results are shown in the following Tables.

\begin{table*}[h]\centering 
\scriptsize
\begin{tabular}{lccccccccc}\toprule
&C-STANCE &FOMC &MeetingBank &ScienceQA &NumGLUE-cm &20Minuten &MedMCQA &JEC-QA \\\cmidrule{2-9}
C-STANCE &\textbf{0.418} &0.323 &0.091 &0.316 &0.099 &0.373 &0.25 &0.28 \\\cmidrule{1-9}
FOMC &0.218 &\textbf{0.603} &0.091 &0.216 &0.049 &0.37 &0.226 &0.256 \\\cmidrule{1-9}
MeetingBank &0.194 &0.494 &\textbf{0.247} &0.048 &0.086 &0.384 &0.244 &0.238 \\\cmidrule{1-9}
ScienceQA &0.32 &0.262 &0.148 &\textbf{0.368} &0 &0.367 &0.226 &0.238 \\\cmidrule{1-9}
NumGLUE-cm &0.256 &0.204 &0.057 &0.244 &\textbf{0.333} &0.37 &0.162 &0.252 \\\cmidrule{1-9}
20Minuten &0.002 &0.012 &0.087 &0.158 &0.148 &\textbf{0.414} &0.222 &0.144 \\\cmidrule{1-9}
MedMCQA &0.326 &0.246 &0.083 &0.292 &0.16 &0.396 &\textbf{0.29} &0.278 \\\cmidrule{1-9}
JEC-QA &0.164 &0.258 &0.093 &0.096 &0.123 &0.39 &0.26 &\textbf{0.296} \\\cmidrule{1-9}
\textbf{Average} &0.237 &0.3 &0.112 &0.217 &0.125 &0.383 &0.235 &0.248 \\\cmidrule{1-9}
\textbf{BWT} &-0.184 &\textbf{AVE} &0.21 \\\midrule
\bottomrule
\end{tabular}
\caption{LoRA adapter sequentially training}\label{tab: }

\end{table*}

\begin{table*}[h]\centering
\scriptsize
\begin{tabular}{lccccccccc}\toprule
&C-STANCE &FOMC &MeetingBank &ScienceQA &NumGLUE-cm &20Minuten &MedMCQA &JEC-QA \\\cmidrule{2-9}
C-STANCE &\textbf{0.408} &0.359 &0.086 &0.18 &0.049 &0.372 &0.254 &0.302 \\\cmidrule{1-9}
FOMC &0.268 &\textbf{0.645} &0.073 &0.09 &0.037 &0.371 &0.238 &0.238 \\\cmidrule{1-9}
MeetingBank &0.294 &0.373 &\textbf{0.252} &0.05 &0.049 &0.378 &0.216 &0.222 \\\cmidrule{1-9}
ScienceQA &0.322 &0.25 &0.122 &0.48 &0.025 &0.375 &0.202 &0.274 \\\cmidrule{1-9}
NumGLUE-cm &0.044 &0.135 &0.048 &0.34 &\textbf{0.37} &0.377 &0.206 &0.244 \\\cmidrule{1-9}
20Minuten &0.112 &0.01 &0.121 &0.154 &0.185 &\textbf{0.411} &0.234 &0.226 \\\cmidrule{1-9}
MedMCQA &0.092 &0.28 &0.111 &0.232 &0.185 &0.392 &\textbf{0.306} &0.228 \\\cmidrule{1-9}
JEC-QA &0.21 &0.292 &0.123 &\textbf{0.238} &0.173 &0.389 &0.262 &\textbf{0.268} \\\cmidrule{1-9}
\textbf{Average} &0.219 &0.293 &0.117 &0.221 &0.134 &0.383 &0.24 &0.25 \\\cmidrule{1-9}
\textbf{BWT} &-0.169 &\textbf{AVE} &0.244  \\\midrule
\bottomrule
\end{tabular}
\caption{ER replay with rate 0.1}\label{tab: }
\end{table*}

\begin{table*}[h]\centering 
\scriptsize
\begin{tabular}{lccccccccc}\toprule
&C-STANCE &FOMC &MeetingBank &ScienceQA &NumGLUE-cm &20Minuten &MedMCQA &JEC-QA \\\cmidrule{2-9}
C-STANCE &\textbf{0.422} &0.488 &0.061 &0.08 &0.086 &0.372 &0.238 &0.248 \\\cmidrule{1-9}
FOMC &0.356 &\textbf{0.706} &0.091 &0.098 &0.062 &0.371 &0.236 &0.232 \\\cmidrule{1-9}
MeetingBank &0.338 &0.484 &\textbf{0.256} &0.082 &0.049 &0.374 &0.24 &0.208 \\\cmidrule{1-9}
ScienceQA &0.374 &0.381 &0.143 &\textbf{0.642} &0.012 &0.371 &0.208 &0.248 \\\cmidrule{1-9}
NumGLUE-cm &0.308 &0.25 &0.068 &0.368 &\textbf{0.284} &0.369 &0.214 &0.288 \\\cmidrule{1-9}
20Minuten &0.19 &0.248 &0.152 &0.424 &0.148 &\textbf{0.415} &0.244 &0.19 \\\cmidrule{1-9}
MedMCQA &0.194 &0.349 &0.086 &0.142 &0.198 &0.401 &\textbf{0.276} &0.142 \\\cmidrule{1-9}
JEC-QA &0.176 &0.29 &0.081 &0.2 &0.148 &0.382 &0.24 &\textbf{0.29} \\\cmidrule{1-9}
\textbf{Average} &0.295 &0.4 &0.117 &0.255 &0.123 &0.382 &0.237 &0.231 \\\cmidrule{1-9}
\textbf{BWT} &-0.212 &\textbf{AVE} &0.226 &\\ \midrule
\bottomrule
\end{tabular}
\caption{LoRA adapter training with EWC}\label{tab: }
\end{table*}

\begin{table*}[!htp]\centering 
\scriptsize
\begin{tabular}{lccccccccc}\toprule
&C-STANCE &FOMC &MeetingBank &ScienceQA &NumGLUE-cm &20Minuten &MedMCQA &JEC-QA \\\cmidrule{2-9}
C-STANCE &\textbf{0.388} &0.29 &0.104 &0.364 &0.062 &0.373 &0.244 &0.268 \\\cmidrule{1-9}
FOMC &0.314 &\textbf{0.615} &0.069 &0.342 &0.074 &0.37 &0.226 &0.26 \\\cmidrule{1-9}
MeetingBank &0.174 &0.444 &\textbf{0.233} &0.182 &0.049 &0.383 &0.228 &0.22 \\\cmidrule{1-9}
ScienceQA &0.318 &0.26 &0.115 &\textbf{0.404} &0.074 &0.373 &0.224 &0.228 \\\cmidrule{1-9}
NumGLUE-cm &0.05 &0.006 &0.074 &0.186 &\textbf{0.321} &0.382 &0.088 &0.158 \\\cmidrule{1-9}
20Minuten &0.052 &0.071 &0.12 &0.24 &0.173 &\textbf{0.409} &0.264 &0.226 \\\cmidrule{1-9}
MedMCQA &0.32 &0.262 &0.077 &0.47 &0.185 &0.389 &\textbf{0.292} &0.25 \\\cmidrule{1-9}
JEC-QA &0.118 &0.296 &0.09 &0.162 &0.136 &0.395 &0.236 &\textbf{0.286} \\\cmidrule{1-9}
\textbf{Average} &0.217 &0.281 &0.11 &0.294 &0.134 &0.384 &0.225 &0.237 \\\cmidrule{1-9}
\textbf{BWT} &-0.176 & \textbf{AVE} &0.215  \\\midrule
\bottomrule
\end{tabular}
\caption{LoRA adapter training with GEM}\label{tab: }
\end{table*}

\begin{table*}[!htp]\centering
\scriptsize
\begin{tabular}{lccccccccc}\toprule
&C-STANCE &FOMC &MeetingBank &ScienceQA &NumGLUE-cm &20Minuten &MedMCQA &JEC-QA \\\cmidrule{2-9}
C-STANCE &\textbf{0.402} &0.28 &0.075 &0.374 &0.062 &0.374 &0.254 &\textbf{0.286} \\\cmidrule{1-9}
FOMC &0.29 &\textbf{0.589} &0.085 &0.26 &0.012 &0.37 &0.266 &0.238 \\\cmidrule{1-9}
MeetingBank &0.118 &0.484 &\textbf{0.24} &0.222 &0.074 &0.38 &0.226 &0.146 \\\cmidrule{1-9}
ScienceQA &0.282 &0.486 &0.171 &\textbf{0.54} &0.062 &0.379 &0.23 &0.258 \\\cmidrule{1-9}
NumGLUE-cm &0.252 &0.183 &0.07 &0.164 &\textbf{0.321} &0.367 &0.138 &0.086 \\\cmidrule{1-9}
20Minuten &0.246 &0.083 &0.081 &0.3 &0.235 &\textbf{0.416} &0.252 &0.16 \\\cmidrule{1-9}
MedMCQA &0.192 &0.462 &0.06 &0.334 &0.21 &0.379 &\textbf{0.278} &0.216 \\\cmidrule{1-9}
JEC-QA &0.316 &0.51 &0.088 &0.35 &0.185 &0.391 &0.278 &0.252 \\\cmidrule{1-9}
\textbf{Average} &0.262 &0.385 &0.109 &0.318 &0.145 &0.382 &0.24 &0.205 \\\cmidrule{1-9}
\textbf{BWT} &-0.095  & \textbf{AVE} &0.296  \\\midrule
\bottomrule
\end{tabular}
\caption{LoRA adapter training with A-GEM}\label{tab: }

\end{table*}

\begin{table*}[!htp]\centering 
\scriptsize
\begin{tabular}{lccccccccc}\toprule
&C-STANCE &FOMC &MeetingBank &ScienceQA &NumGLUE-cm &20Minuten &MedMCQA &JEC-QA \\\cmidrule{2-9}
C-STANCE &\textbf{0.438} &0.258 &0.115 &0.42 &0.062 &0 &0 &0 \\\cmidrule{1-9}
FOMC &0.26 &\textbf{0.53} &0.079 &0.22 &0.074 &0.372 &0.226 &0.234 \\\cmidrule{1-9}
MeetingBank &0 &0.472 &\textbf{0.248} &0.022 &0.025 &0.374 &0.176 &0.152 \\\cmidrule{1-9}
ScienceQA &0.324 &0.254 &0.166 &0.312 &0.062 &0.382 &0.2 &\textbf{0.278} \\\cmidrule{1-9}
NumGLUE-cm &0.194 &0.363 &0.045 &0.254 &\textbf{0.284} &0.375 &0.158 &0.27 \\\cmidrule{1-9}
20Minuten &0 &0.002 &0.128 &0.16 &0.235 &\textbf{0.402} &\textbf{0.288} &0.198 \\\cmidrule{1-9}
MedMCQA &0.002 &0.264 &0.143 &\textbf{0.408} &0.259 &0.393 &0.258 &0.262 \\\cmidrule{1-9}
JEC-QA &0.18 &0.335 &0.128 &0.322 &0.173 &0.391 &0.258 &0.268 \\\cmidrule{1-9}
\textbf{Average} &0.175 &0.31 &0.132 &0.265 &0.147 &0.336 &0.196 &0.208 \\\cmidrule{1-9}
\textbf{BWT} &-0.098 &\textbf{AVE} &0.257  \\\midrule
\bottomrule
\end{tabular}
\caption{LoRA adapter training with Learning to Prompt}\label{tab: }
\end{table*}

\begin{table*}[!htp]\centering 
\scriptsize
\begin{tabular}{lccccccccc}\toprule
&C-STANCE &FOMC &MeetingBank &ScienceQA &NumGLUE-cm &20Minuten &MedMCQA &JEC-QA \\\cmidrule{2-9}
C-STANCE &\textbf{0.372} &0.244 &0.073 &0.008 &0.012 &0.374 &0.08 &0.108 \\\cmidrule{1-9}
FOMC &0.324 &\textbf{0.518} &0.057 &0.02 &0.025 &0.373 &0.158 &0.236 \\\cmidrule{1-9}
MeetingBank &0.332 &0.254 &\textbf{0.211} &0.032 &0 &0.374 &0.212 &0.048 \\\cmidrule{1-9}
ScienceQA &0.324 &0.242 &0.147 &0.422 &0.025 &0.377 &0.23 &0.248 \\\cmidrule{1-9}
NumGLUE-cm &0.164 &0.49 &0.079 &0.262 &\textbf{0.272} &0.378 &0.172 &0.18 \\\cmidrule{1-9}
20Minuten &0.32 &0.131 &0.093 &0.408 &0.198 &\textbf{0.406} &0.218 &0.254 \\\cmidrule{1-9}
MedMCQA &0.244 &0.341 &0.064 &\textbf{0.424} &0.111 &0.387 &\textbf{0.268} &0.224 \\\cmidrule{1-9}
JEC-QA &0.024 &0.476 &0.059 &0.226 &0.086 &0.381 &0.226 &\textbf{0.312} \\\cmidrule{1-9}
\textbf{Average} &0.263 &0.337 &0.098 &0.225 &0.091 &0.381 &0.196 &0.201 \\\cmidrule{1-9}
\textbf{BWT} &-0.142 &\textbf{AVE} &0.224  \\\midrule
\bottomrule
\end{tabular}
\caption{LoRA training with Progerssize Prompts\label{tab: pp}}
\end{table*}

\begin{table*}[!htp]\centering
\scriptsize
\begin{tabular}{lccccccccc}\toprule
C-STANCE &FOMC &MeetingBank &ScienceQA &NumGLUE-cm &20Minuten &MedMCQA &JEC-QA \\\cmidrule{1-8}
0.384 &0.506 &0.252 &0.586 &0.284 &0.416 &0.248 &0.244 \\\midrule
\bottomrule
\end{tabular}
\caption{Multi task training}\label{tab: }
\end{table*}

\begin{table*}[!htp]\centering
\scriptsize
\begin{tabular}{lccccccccc}\toprule
C-STANCE &FOMC &MeetingBank &ScienceQA &NumGLUE-cm &20Minuten &MedMCQA &JEC-QA \\\cmidrule{1-8}
0.35 &0.24 &0.10 &0.23 &0.28 &0.30 &0.24 &0.12 \\\midrule
\bottomrule
\end{tabular}
\caption{zero-shot inference results}\label{tab: zero-shot}
\end{table*}

\begin{table*}[!htp]\centering
\scriptsize
\begin{tabular}{lccccccccc}\toprule
&C-STANCE &FOMC &MeetingBank &ScienceQA &NumGLUE-cm &20Minuten &MedMCQA &JEC-QA \\\cmidrule{2-9}
C-STANCE &\textbf{0.388} &0.276 &0.081 &0.426 &0.049 &0.377 &0.226 &0.278 \\\cmidrule{1-9}
FOMC &0.342 &\textbf{0.488} &0.101 &0.074 &0.049 &0.373 &0.23 &0.202 \\\cmidrule{1-9}
MeetingBank &0.164 &0.032 &\textbf{0.233} &0.002 &0.037 &0.388 &0.222 &0.194 \\\cmidrule{1-9}
ScienceQA &0.324 &0.252 &0.099 &\textbf{0.434} &0.012 &0.382 &0.196 &0.282 \\\cmidrule{1-9}
NumGLUE-cm &0.308 &0.004 &0.059 &0.31 &\textbf{0.148} &0.375 &0.2 &0.25 \\\cmidrule{1-9}
20Minuten &0.014 &0.014 &0.063 &0.032 &0.062 &\textbf{0.412} &0.07 &0.018 \\\cmidrule{1-9}
MedMCQA &0.328 &0.236 &0.057 &0.062 &0.111 &0.373 &\textbf{0.258} &0.274 \\\cmidrule{1-9}
JEC-QA &0.324 &0.272 &0.068 &0.222 &0.049 &0.381 &0.25 &\textbf{0.284} \\\midrule
\bottomrule
\end{tabular}
\caption{Single task fine-tuning}\label{tab: }
\end{table*}

\begin{table*}[!htp]\centering
\caption{Linear combination with learning fast and slow}\label{tab: }
\scriptsize
\begin{tabular}{lccccccccc}\toprule
&C-STANCE &FOMC &MeetingBank &ScienceQA &NumGLUE-cm &20Minuten &MedMCQA &JEC-QA \\\cmidrule{2-9}
C-STANCE &\textbf{}{0.444} &0.357 &0.066 &0.122 &0.111 &0.374 &0.254 &\textbf{0.292} \\\cmidrule{1-9}
FOMC &0.432 &\textbf{0.645} &0.066 &0.1 &0.074 &0.37 &0.236 &0.266 \\\cmidrule{1-9}
MeetingBank &0.184 &0.522 &0.213 &0.26 &0.074 &0.384 &0.244 &0.236 \\\cmidrule{1-9}
ScienceQA &0.354 &0.51 &0.192 &\textbf{0.542} &0.025 &0.378 &0.214 &0.226 \\\cmidrule{1-9}
NumGLUE-cm &0.358 &0.438 &0.147 &0.448 &\textbf{0.296} &0.383 &0.196 &0.268 \\\cmidrule{1-9}
20Minuten &0.336 &0.081 &0.144 &0.442 &0.222 &\textbf{0.414} &0.244 &0.206 \\\cmidrule{1-9}
MedMCQA &0.43 &0.611 &0.213 &0.496 &0.235 &0.411 &0.27 &0.29 \\\cmidrule{1-9}
JEC-QA &0.43 &0.601 &\textbf{0.216} &0.486 &0.222 &0.402 &\textbf{0.276} &0.272 \\\cmidrule{1-9}
\textbf{Average} &0.371 &0.471 &0.157 &0.362 &0.157 &0.39 &0.242 &0.257 \\\cmidrule{1-9}
\textbf{BWT} &-0.027 &AVE &0.363 \\\midrule
\bottomrule
\end{tabular}
\end{table*}

\onecolumn
\begin{table*}[!htp]
\centering
\caption{Detailed Results of General Benchmarks}\label{tab:DetailResultsGeneralBenchmarks}
\footnotesize
\resizebox{\textwidth}{!}{
\begin{tabular}{lrrrrrrrrrrrrr}
\toprule
dataset &version &metric &mode &SEQ &EWC &ER &GEM &AGEM &L2P &PP &MTL &OURS \\\cmidrule{1-13}
lukaemon\_mmlu\_college\_biology &8c2e29 &accuracy &gen &0.69 &20.83 &0.69 &14.58 &2.78 &3.47 &33.33 &21.53 &0 \\\cmidrule{1-13}
lukaemon\_mmlu\_college\_chemistry &0afccd &accuracy &gen &2 &12 &3 &7 &3 &1 &13 &13 &3 \\\cmidrule{1-13}
lukaemon\_mmlu\_college\_computer\_science &c1c1b4 &accuracy &gen &5 &16 &7 &18 &9 &3 &22 &8 &4 \\\cmidrule{1-13}
lukaemon\_mmlu\_college\_mathematics &9deed0 &accuracy &gen &8 &23 &11 &7 &2 &0 &13 &18 &4 \\\cmidrule{1-13}
lukaemon\_mmlu\_college\_physics &f5cf5e &accuracy &gen &0 &8.82 &0 &2.94 &0 &0.98 &3.92 &15.69 &0 \\\cmidrule{1-13}
lukaemon\_mmlu\_electrical\_engineering &3d694d &accuracy &gen &0 &13.79 &8.28 &6.21 &1.38 &1.38 &31.03 &20 &1.38 \\\cmidrule{1-13}
lukaemon\_mmlu\_astronomy &7ef16f &accuracy &gen &4.61 &12.5 &1.97 &33.55 &1.32 &0 &26.97 &16.45 &6.58 \\\cmidrule{1-13}
lukaemon\_mmlu\_anatomy &2d597d &accuracy &gen &1.48 &17.04 &0 &9.63 &5.93 &0 &37.04 &5.93 &5.19 \\\cmidrule{1-13}
lukaemon\_mmlu\_abstract\_algebra &ec092c &accuracy &gen &7 &25 &8 &15 &2 &3 &22 &23 &12 \\\cmidrule{1-13}
lukaemon\_mmlu\_machine\_learning &d489ae &accuracy &gen &17.86 &25 &14.29 &3.57 &0 &0.89 &20.54 &4.46 &0 \\\cmidrule{1-13}
lukaemon\_mmlu\_clinical\_knowledge &af10df &accuracy &gen &0.75 &26.42 &1.89 &8.68 &0.75 &0 &34.34 &10.19 &1.51 \\\cmidrule{1-13}
lukaemon\_mmlu\_global\_facts &cad9e0 &accuracy &gen &1 &29 &10 &30 &3 &0 &22 &28 &5 \\\cmidrule{1-13}
lukaemon\_mmlu\_management &65f310 &accuracy &gen &0 &25.24 &1.94 &24.27 &0 &0 &30.1 &14.56 &0 \\\cmidrule{1-13}
lukaemon\_mmlu\_nutrition &80bf96 &accuracy &gen &0 &19.61 &2.29 &15.36 &7.19 &0.65 &32.03 &8.82 &11.11 \\\cmidrule{1-13}
lukaemon\_mmlu\_marketing &9a98c0 &accuracy &gen &0.43 &25.64 &2.56 &5.13 &11.54 &21.37 &50.85 &7.26 &0.43 \\\cmidrule{1-13}
lukaemon\_mmlu\_professional\_accounting &9cc7e2 &accuracy &gen &5.32 &20.21 &15.96 &21.99 &12.06 &13.83 &20.92 &20.92 &13.83 \\\cmidrule{1-13}
lukaemon\_mmlu\_high\_school\_geography &c28a4c &accuracy &gen &0.51 &21.72 &0.51 &5.56 &1.52 &0 &33.84 &2.53 &5.56 \\\cmidrule{1-13}
lukaemon\_mmlu\_international\_law &408d4e &accuracy &gen &32.23 &24.79 &1.65 &52.07 &47.93 &1.65 &47.93 &45.45 &0 \\\cmidrule{1-13}
lukaemon\_mmlu\_moral\_scenarios &9f30a6 &accuracy &gen &0 &24.25 &0.11 &19.33 &24.25 &0 &24.13 &24.25 &0.22 \\\cmidrule{1-13}
lukaemon\_mmlu\_computer\_security &2753c1 &accuracy &gen &1 &18 &2 &4 &0 &0 &44 &17 &0 \\\cmidrule{1-13}
lukaemon\_mmlu\_high\_school\_microeconomics &af9eae &accuracy &gen &0.42 &22.69 &0 &21.85 &12.18 &0.42 &27.73 &9.24 &13.87 \\\cmidrule{1-13}
lukaemon\_mmlu\_professional\_law &7c7a62 &accuracy &gen &6.06 &18.9 &3.65 &29.14 &16.36 &0.52 &29.14 &21.71 &3.13 \\\cmidrule{1-13}
lukaemon\_mmlu\_medical\_genetics &b1a3a7 &accuracy &gen &0 &19 &5 &3 &1 &1 &35 &5 &0 \\\cmidrule{1-13}
lukaemon\_mmlu\_professional\_psychology &c6b790 &accuracy &gen &0.98 &14.38 &1.31 &22.88 &11.44 &0.16 &39.22 &13.24 &0.98 \\\cmidrule{1-13}
lukaemon\_mmlu\_jurisprudence &f41074 &accuracy &gen &0 &28.7 &0 &29.63 &0.93 &0 &43.52 &25 &0 \\\cmidrule{1-13}
lukaemon\_mmlu\_world\_religions &d44a95 &accuracy &gen &0.58 &20.47 &4.68 &41.52 &5.85 &4.09 &47.37 &2.92 &1.17 \\\cmidrule{1-13}
lukaemon\_mmlu\_philosophy &d36ef3 &accuracy &gen &1.61 &27.01 &3.54 &4.5 &11.9 &1.61 &37.94 &7.4 &0.96 \\\cmidrule{1-13}
lukaemon\_mmlu\_virology &0a5f8e &accuracy &gen &0 &29.52 &9.04 &3.01 &0.6 &0 &33.13 &16.87 &2.41 \\\cmidrule{1-13}
lukaemon\_mmlu\_high\_school\_chemistry &5b2ef9 &accuracy &gen &2.46 &15.76 &2.46 &10.84 &3.45 &1.48 &29.06 &14.29 &2.46 \\\cmidrule{1-13}
lukaemon\_mmlu\_public\_relations &4c7898 &accuracy &gen &0.91 &31.82 &11.82 &4.55 &0.91 &0 &32.73 &20 &0 \\\cmidrule{1-13}
lukaemon\_mmlu\_high\_school\_macroeconomics &3f841b &accuracy &gen &4.87 &19.74 &8.46 &13.08 &2.56 &0 &22.56 &7.95 &7.18 \\\cmidrule{1-13}
lukaemon\_mmlu\_human\_sexuality &4d1f3e &accuracy &gen &0.76 &15.27 &4.58 &2.29 &2.29 &0 &35.88 &4.58 &0.76 \\\cmidrule{1-13}
lukaemon\_mmlu\_elementary\_mathematics &0f5d3a &accuracy &gen &1.32 &14.29 &5.03 &21.96 &3.44 &2.12 &18.25 &12.96 &7.14 \\\cmidrule{1-13}
lukaemon\_mmlu\_high\_school\_physics &0dd929 &accuracy &gen &5.96 &12.58 &1.99 &13.25 &3.97 &2.65 &20.53 &15.89 &8.61 \\\cmidrule{1-13}
lukaemon\_mmlu\_high\_school\_computer\_science &bf31fd &accuracy &gen &5 &23 &8 &17 &4 &1 &30 &18 &3 \\\cmidrule{1-13}
lukaemon\_mmlu\_high\_school\_european\_history &d1b67e &accuracy &gen &16.97 &23.03 &12.12 &21.21 &15.76 &11.52 &26.67 &9.7 &6.06 \\\cmidrule{1-13}
lukaemon\_mmlu\_business\_ethics &af53f3 &accuracy &gen &0 &26 &0 &5 &1 &1 &39 &15 &0 \\\cmidrule{1-13}
lukaemon\_mmlu\_moral\_disputes &48239e &accuracy &gen &0 &24.86 &0.58 &24.57 &4.05 &8.09 &37.28 &5.78 &9.54 \\\cmidrule{1-13}
lukaemon\_mmlu\_high\_school\_statistics &47e18e &accuracy &gen &4.63 &14.81 &8.8 &18.52 &5.09 &2.78 &15.74 &3.7 &7.87 \\\cmidrule{1-13}
lukaemon\_mmlu\_miscellaneous &573569 &accuracy &gen &1.4 &27.71 &19.67 &23.5 &5.49 &0.51 &44.7 &24.14 &7.41 \\\cmidrule{1-13}
lukaemon\_mmlu\_formal\_logic &7a0414 &accuracy &gen &2.38 &18.25 &7.14 &17.46 &15.87 &1.59 &19.84 &10.32 &6.35 \\\cmidrule{1-13}
lukaemon\_mmlu\_high\_school\_government\_and\_politics &d907eb &accuracy &gen &0 &20.73 &0.52 &30.57 &2.59 &0 &36.27 &22.28 &17.62 \\\cmidrule{1-13}
lukaemon\_mmlu\_prehistory &65aa94 &accuracy &gen &3.4 &23.15 &4.01 &18.83 &4.32 &1.23 &36.73 &21.3 &1.85 \\\cmidrule{1-13}
lukaemon\_mmlu\_security\_studies &9ea7d3 &accuracy &gen &0.41 &15.92 &2.04 &26.53 &6.53 &3.27 &26.53 &16.73 &3.67 \\\cmidrule{1-13}
lukaemon\_mmlu\_high\_school\_biology &775183 &accuracy &gen &0.65 &25.16 &0.65 &9.03 &3.23 &0.97 &37.1 &1.94 &14.19 \\\cmidrule{1-13}
lukaemon\_mmlu\_logical\_fallacies &19746a &accuracy &gen &1.84 &24.54 &6.75 &12.88 &12.88 &0.61 &31.9 &11.04 &6.13 \\\cmidrule{1-13}
lukaemon\_mmlu\_high\_school\_world\_history &6665dc &accuracy &gen &18.57 &26.58 &10.13 &7.59 &23.63 &23.21 &21.1 &9.7 &8.44 \\\cmidrule{1-13}
lukaemon\_mmlu\_professional\_medicine &a05bab &accuracy &gen &9.93 &15.07 &4.04 &8.09 &1.1 &0 &38.97 &1.1 &5.51 \\\cmidrule{1-13}
lukaemon\_mmlu\_high\_school\_mathematics &0e6a7e &accuracy &gen &4.81 &16.3 &5.56 &22.59 &3.33 &0.37 &18.89 &13.7 &3.33 \\\cmidrule{1-13}
lukaemon\_mmlu\_college\_medicine &5215f1 &accuracy &gen &1.16 &14.45 &2.31 &5.78 &1.16 &1.73 &28.9 &9.25 &2.31 \\\cmidrule{1-13}
lukaemon\_mmlu\_high\_school\_us\_history &b5f235 &accuracy &gen &8.82 &18.63 &1.47 &20.1 &11.76 &2.94 &18.63 &9.31 &2.94 \\\cmidrule{1-13}
lukaemon\_mmlu\_sociology &4980ec &accuracy &gen &3.98 &21.89 &8.96 &4.48 &5.97 &0 &41.79 &6.47 &2.49 \\\cmidrule{1-13}
lukaemon\_mmlu\_econometrics &4d590b &accuracy &gen &7.02 &24.56 &2.63 &16.67 &12.28 &0.88 &25.44 &9.65 &0.88 \\\cmidrule{1-13}
lukaemon\_mmlu\_high\_school\_psychology &440e98 &accuracy &gen &0.73 &22.57 &0.55 &10.28 &8.99 &0.73 &37.61 &9.72 &12.48 \\\cmidrule{1-13}
lukaemon\_mmlu\_human\_aging &d0a8e1 &accuracy &gen &0.9 &36.77 &15.25 &0.45 &0.9 &0 &39.01 &23.32 &4.93 \\\cmidrule{1-13}
lukaemon\_mmlu\_us\_foreign\_policy &adcc88 &accuracy &gen &1 &21 &13 &22 &5 &0 &44 &14 &20 \\\cmidrule{1-13}
lukaemon\_mmlu\_conceptual\_physics &a111d3 &accuracy &gen &0 &22.13 &8.51 &11.91 &0.85 &0 &31.91 &28.09 &2.55 \\\cmidrule{1-13}
\bottomrule
\end{tabular}}
\end{table*}

\onecolumn
\begin{table*}[!htp]
\centering
\caption{Detailed Results of General Benchmarks}\label{tab:DetailResultsGeneralBenchmarks}
\footnotesize
\resizebox{\textwidth}{!}{
\begin{tabular}{lrrrrrrrrrrrrr}
\toprule
dataset &version &metric &mode &SEQ &EWC &ER &GEM &AGEM &L2P &PP &MTL &OURS \\\cmidrule{1-13}
bbh-temporal\_sequences &e43931 &score &gen &8.4 &16.4 &19.2 &17.2 &18.4 &22.4 &24.4 &19.6 &15.2 \\\cmidrule{1-13}
bbh-disambiguation\_qa &d52c61 &score &gen &30 &30.8 &31.2 &29.6 &30 &30 &30 &34.8 &30.8 \\\cmidrule{1-13}
bbh-date\_understanding &a8000b &score &gen &32 &26 &27.6 &36.8 &32 &39.2 &33.2 &38 &27.6 \\\cmidrule{1-13}
bbh-tracking\_shuffled\_objects\_three\_objects &7964c0 &score &gen &32.8 &34.4 &29.2 &28.4 &33.2 &31.6 &29.2 &35.2 &32.8 \\\cmidrule{1-13}
bbh-penguins\_in\_a\_table &fceb27 &score &gen &32.88 &29.45 &28.08 &26.03 &35.62 &35.62 &30.14 &32.88 &30.82 \\\cmidrule{1-13}
bbh-geometric\_shapes &503c8f &score &gen &4.8 &8.8 &0.4 &0.8 &0 &0 &3.2 &2 &3.2 \\\cmidrule{1-13}
bbh-snarks &42d6ca &score &gen &46.07 &50 &47.75 &54.49 &53.37 &50.56 &51.12 &50 &48.31 \\\cmidrule{1-13}
bbh-ruin\_names &408de8 &score &gen &23.2 &27.2 &24.4 &24.8 &22.4 &24.4 &22.4 &29.2 &24 \\\cmidrule{1-13}
bbh-tracking\_shuffled\_objects\_seven\_objects &7964c0 &score &gen &17.6 &19.6 &16 &16 &17.6 &13.6 &17.2 &12.4 &15.6 \\\cmidrule{1-13}
bbh-tracking\_shuffled\_objects\_five\_objects &7964c0 &score &gen &16.8 &17.6 &14 &16 &20.8 &13.2 &17.2 &14.4 &18.4 \\\cmidrule{1-13}
bbh-logical\_deduction\_three\_objects &45ebc5 &score &gen &35.6 &32.4 &42 &44 &50.4 &45.6 &42 &41.2 &39.6 \\\cmidrule{1-13}
bbh-hyperbaton &5e5016 &score &gen &53.2 &55.2 &54.8 &53.6 &55.6 &56.8 &53.6 &53.2 &48.8 \\\cmidrule{1-13}
bbh-logical\_deduction\_five\_objects &45ebc5 &score &gen &22.8 &17.2 &19.6 &21.2 &28.8 &23.2 &23.2 &22.8 &27.6 \\\cmidrule{1-13}
bbh-logical\_deduction\_seven\_objects &45ebc5 &score &gen &14.4 &12.4 &10.4 &19.2 &20.4 &19.2 &19.6 &13.6 &13.2 \\\cmidrule{1-13}
bbh-movie\_recommendation &cc2fde &score &gen &31.2 &41.2 &36.4 &62.4 &60.8 &63.6 &62.4 &53.6 &34 \\\cmidrule{1-13}
bbh-salient\_translation\_error\_detection &5b5f35 &score &gen &10.8 &11.6 &13.6 &14 &18 &11.2 &12 &18.8 &6.4 \\\cmidrule{1-13}
bbh-reasoning\_about\_colored\_objects &1cb761 &score &gen &21.6 &16.8 &18.4 &22.4 &25.2 &26 &24.4 &22.8 &21.6 \\\cmidrule{1-13}
bbh-multistep\_arithmetic\_two &30f91e &score &gen &0 &0 &1.2 &0.4 &0.4 &1.2 &0.4 &1.6 &1.2 \\\cmidrule{1-13}
bbh-navigate &1576d9 &score &gen &40.8 &43.2 &51.6 &45.6 &46.8 &48.8 &48 &54.8 &46 \\\cmidrule{1-13}
bbh-dyck\_languages &805bea &score &gen &0 &0.4 &0 &0.4 &0 &0 &0.8 &0 &0.4 \\\cmidrule{1-13}
bbh-word\_sorting &9a3f78 &score &gen &3.2 &1.2 &1.6 &5.2 &4.8 &4.8 &6.4 &6.8 &2.8 \\\cmidrule{1-13}
bbh-sports\_understanding &d3fa77 &score &gen &87.2 &79.2 &90.8 &88 &78.8 &84 &84 &77.6 &90.8 \\\cmidrule{1-13}
bbh-boolean\_expressions &612c92 &score &gen &61.6 &47.2 &65.2 &62.8 &63.6 &60.4 &55.6 &61.2 &63.6 \\\cmidrule{1-13}
bbh-object\_counting &781e5c &score &gen &47.6 &50.8 &41.6 &45.6 &50 &50.4 &56 &48 &43.2 \\\cmidrule{1-13}
bbh-formal\_fallacies &eada96 &score &gen &15.6 &12.8 &2.8 &26.8 &24 &19.6 &27.2 &13.6 &2.4 \\\cmidrule{1-13}
bbh-causal\_judgement &89eaa4 &score &gen &36.9 &34.76 &37.43 &41.71 &35.83 &39.57 &44.92 &50.27 &33.16 \\\cmidrule{1-13}
bbh-web\_of\_lies &0c0441 &score &gen &51.2 &42.8 &51.2 &53.6 &49.2 &47.6 &53.6 &53.6 &52.4 \\\cmidrule{1-13}
piqa &1194eb &accuracy &gen &58.49 &47.12 &53.1 &53.48 &53.92 &54.19 &53.05 &52.99 &51.25 \\\midrule
\bottomrule
\end{tabular}}
\end{table*}

\end{document}